\documentclass[conference]{IEEEtran}
\IEEEoverridecommandlockouts
\usepackage{cite}
\usepackage{amsmath,amssymb,amsfonts}
\usepackage{algorithmic}
\usepackage{graphicx}
\usepackage{textcomp}
\usepackage{xcolor}
\usepackage{booktabs}
\usepackage{float}
\usepackage{booktabs}
\usepackage{tabularx}
\usepackage{multirow}
\usepackage{cite}
\usepackage{multirow}
\usepackage{multicol}
\usepackage{tabularx}
\usepackage{colortbl}
\newcolumntype{Y}{>{\centering\arraybackslash}X}
\def\BibTeX{{\rm B\kern-.05em{\sc i\kern-.025em b}\kern-.08em
    T\kern-.1667em\lower.7ex\hbox{E}\kern-.125emX}}
    
\title{DocVideoQA: Towards Comprehensive Understanding of Document-Centric Videos through Question Answering}

\author{
    \IEEEauthorblockN{Haochen Wang}
    \IEEEauthorblockA{\textit{Peking University} \\ Beijing, China \\ wanghaochen326@stu.pku.edu.cn}
    \and
    \IEEEauthorblockN{Kai Hu}
    \IEEEauthorblockA{\textit{University of Science and Technology of China} \\ Hefei, China \\ hk970213@mail.ustc.edu.cn}
    \and
    \IEEEauthorblockN{Liangcai Gao\thanks{\textsuperscript{*}Corresponding Author}}
    \IEEEauthorblockA{\textit{Peking University} \\ Beijing, China \\ gaoliangcai@pku.edu.cn}
}

\begin{document}
\maketitle

\begin{abstract}
Remote work and online courses have become important methods of knowledge dissemination, leading to a large number of document-based instructional videos. Unlike traditional video datasets, these videos mainly feature rich-text images and audio that are densely packed with information closely tied to the visual content, requiring advanced multimodal understanding capabilities. However, this domain remains underexplored due to dataset availability and its inherent complexity. In this paper, we introduce the DocVideoQA task and dataset for the first time, comprising 1,454 videos across 23 categories with a total duration of about 828 hours. The dataset is annotated with 154K question-answer pairs generated manually and via GPT, assessing models' comprehension, temporal awareness, and modality integration capabilities. Initially, we establish a baseline using open-source MLLMs. Recognizing the challenges in modality comprehension for document-centric videos, we present DV-LLaMA, a robust video MLLM baseline. Our method enhances unimodal feature extraction with diverse instruction-tuning data and employs contrastive learning to strengthen modality integration. Through fine-tuning, the LLM is equipped with audio-visual capabilities, leading to significant improvements in document-centric video understanding. Extensive testing on the DocVideoQA dataset shows that DV-LLaMA significantly outperforms existing models. We’ll release the code and dataset to facilitate future research.

\end{abstract}


\begin{figure*}[t]
    \centering
    \includegraphics[width=1\textwidth, height=0.35\textheight,  keepaspectratio]{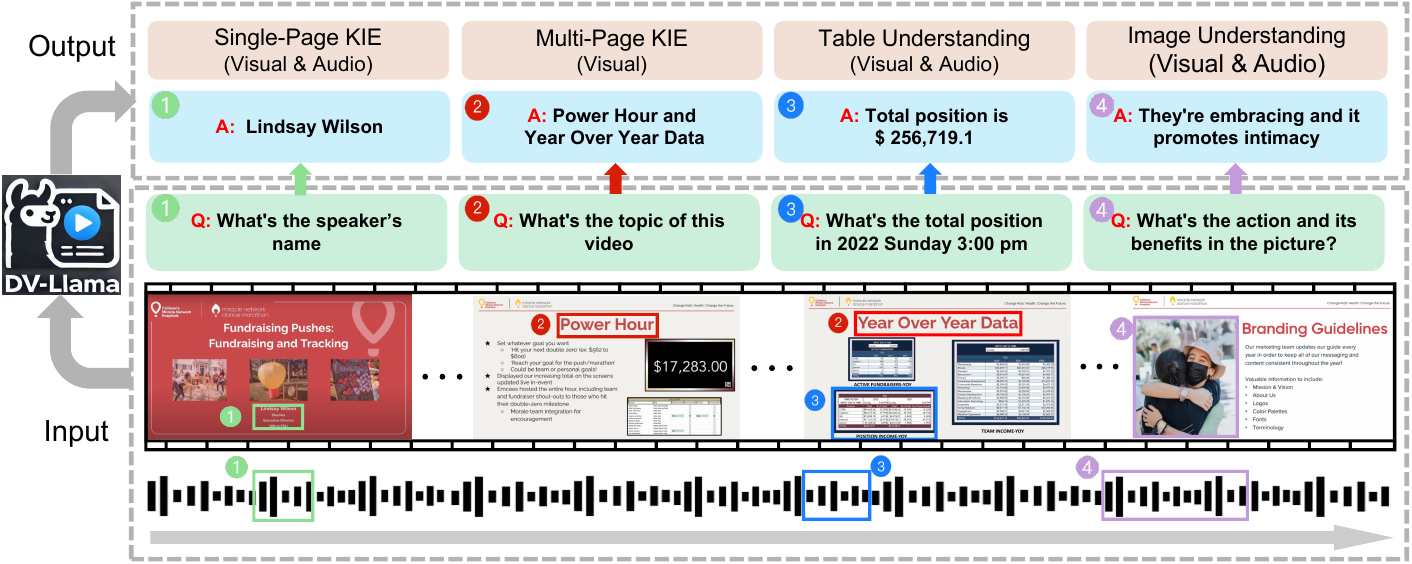}
    \caption{An illustrative example of DocVideoQA: a video on Fundraising Pushes challenges models with tasks such as information extraction, multi-page content comprehension, and visual-audio understanding, necessitating a multidimensional analysis of the video.}
    \label{task_pic}
\end{figure*}

\section{Introduction}
The task of question answering (QA) has long served as a fundamental way for assessing the capacity of machines to understand content across different types. As its core, QA systems aim to automatically answer questions posed by humans, by identifying and extracting relevant information from various media such as text, images, and videos. With advancements in technology, the field of QA has evolved from addressing questions in a single modality to tackling multi-modal questions, including Video Question Answering (VideoQA\cite{b1}). Additionally, the scope of QA has evolved from simple factoid-based questions\cite{b2} to more complex, context-dependent questions\cite{b3} that require deep understanding and reasoning over extensive documents. The rise of online education and remote work has led to a rapid increase in document-centric instructional videos on the internet, making these videos significant medium for knowledge dissemination. However, the investigation of this video category remains constrained due to the unavailability of specialized datasets. Consequently, as illustrated in Fig~\ref{task_pic}, building upon the foundational concept of DocVideoQA, we introduce a new dataset to facilitate the exploration of this novel QA paradigm.

While QA systems have greatly diversified in terms of content input, ranging from short-text snippets \cite{b4}\cite{b5}\cite{b6} to long documents \cite{b7}, and from purely textual data to mixed media such as images (VQA \cite{b8}, ChartQA \cite{b9}, TextVQA \cite{b10}), document images (DocVQA \cite{b11}) and videos (VideoQA \cite{b1}), current research has not adequately focused on the task of Document Video QA. This task, closely related to VideoQA and DocumentQA, encounters unique challenges. In the realm of VideoQA, most studies focus on aspects such as action recognition, sentiment analysis, or event detection, primarily analyzing dynamic visual elements in natural scene videos.In contrast, document videos have distinct characteristics. They contain rich text and intricate layout relationships, as well as audio content covering a wide range of topics and specialized terminology, especially when closely linked to visual information. Therefore, models need to possess advanced capabilities in text and layout analysis, as well as the ability to interpret complex audio-visual interactions. On the other hand, DocumentQA research has progressed from basic text extraction using OCR to the more complex DocVQA task, which combines document layout analysis with visual modalities. However, the field still lacks a task specifically focused on examining document content embedded within the intricate medium of video. The incorporation of document content in videos, along with the integration of temporal dimensions and speech modalities, presents novel challenges in effectively extracting and synthesizing features across different modalities for comprehensive understanding.

In various scenarios, a pronounced need emerges for the capacity to interpret document videos. Within the educational domain, this facilitates the creation of customized review assistance systems tailored to online courses; in the realm of remote work, it enables the swift comprehension of meeting highlights, augmentation of omitted details, and summarization of discussions. To advance this field of research and tackle the previously mentioned challenges, we introduce a novel DocVideoQA dataset to alleviate the issue of data scarcity. Given that slide decks are one of the most efficient document types that arrange visual and textual elements for communication, our dataset comprises 1,454 explanatory videos primarily featuring slides, covering 23 distinct fields. We annotated these videos with 154K pairs of questions and answers, utilizing both human and GPT-generated methods. Our question design spans multiple dimensions (specific dimensions) to comprehensively assess models' understanding of video content. Furthermore, to more effectively extract and integrate features from visual and speech modalities, we developed a robust multi-modal language model benchmark—DV-LLaMA. By employing a progressive training paradigm, our model effectively aligns video content with language, enhancing comprehension abilities by approximately 20\%. Extensive experiments conducted on the DocVideoQA dataset demonstrate that our model achieves the best results in the proposed task. 

The main contributions of this paper can be summarized as follows:
\begin{itemize}
  \item We are the first to present the Document Video Question Answering task, designed to answer questions derived from Document-Centric Video.
  \item To support research, we introduce a new dataset, which consists of 1,454 document-centric videos across 23 domains, annotated with 154K question-answer pairs. We’ll release the code and dataset to facilitate future research.
  \item We developed DV-LLaMA, a Multi-Model Large Language Model, enhancing understanding of Document-Centric Video through improved feature extraction and integration from visual and audio modalities via progressive training.
\end{itemize}

\section{DocVideoQA Dataset Construction}
\begin{table}[t]
\centering
\caption{\textbf{Comparison of VideoQA Datasets} \\ "MC/OE": "multichoice/open-ended". "F/S": "full/segmental".}
\renewcommand{\arraystretch}{1.2}
\footnotesize
\setlength{\tabcolsep}{1.9pt}
\begin{tabularx}{1\columnwidth}{cYYYYYY}
\toprule
Dataset & Topic & Video\_n & QA\_n & Annotation & Task & Duration(s) \\
\midrule
MSVD-QA \cite{b12} & General & 2K & 51K & Auto & OE & 10 \\
NExt-QA \cite{b13} & General & 5.4K & 52K & Man & OE\&MC & 44 \\
TGIF-QA \cite{b14} & Gif & 72K & 165K & Auto\&Man & OE\&MC & 3 \\
MovieQA \cite{b15} & Movie & 408 & 15K & Man & MC & 203 \\
Social-IQA \cite{b16} & Social & 1K & 8K & Man & MC & - \\
DramaQA \cite{b17} & Drama & 24K & 16K & Auto\&Man & MC & 3.7 \\
Sports-QA \cite{b18} & Sports & 6K & 94K & Auto\&Man & OE & 21 \\
\midrule
\multirow{2}{*}{\textbf{DocVideoQA}}& \multirow{2}{*}{Document} & 1.5K$^{F}$ & \multirow{2}{*}{154K} & \multirow{2}{*}{Auto\&Man} & \multirow{2}{*}{OE} & 1,380$^{F}$ \\
& & 75K$^{S}$ & & & & 35$^{S}$ \\
\bottomrule
\end{tabularx}
\label{tab:data}
\end{table}

We built the DocVideoQA dataset from two main sources: the SlideSpeech dataset\cite{b20}, a large-scale audio-visual corpus enriched with slides from multiple domains, and videos from SlidesLive\footnote{https://slideslive.com/}, which consist of academic conference presentations characterized by their high level of professionalism and extensive use of domain-specific vocabulary. From these sources, we selected videos in English that predominantly featured slides for our dataset. For supervised model training and generation of question-answer pairs, we meticulously annotated the video content. Visually, following the SlideSpeech paper, we annotated the timestamps for each slide in the video and utilized the TD\cite{b35} and OCR\cite{b36} models from the MMOCR\cite{b37} toolkit to extract OCR results and restore the correct reading sequence of the text for each slide. For audio, we utilized Whisper\cite{b21} and manual correction to transcribe the audio content of the video into high-confidence text.

In the question-answer pair annotation process, we classified questions into three categories: \textbf{Information extraction}, \textbf{Content comprehension}, and \textbf{Temporal awareness}. This classification requires the model to comprehend visual content, including text, tables, and images, integrate audio information for multimodal understanding, and address cross-page temporal challenges. Some question-answer pairs were generated using GPT-4\cite{b32} in conjunction with the annotated visual and audio text. Additionally, other pairs were manually annotated, particularly focusing on visual content within the video, such as images and tables, that OCR could not capture.

As depicted in Table~\ref{tab:data}, we annotated 1,454 videos and 154K question-answer pairs across 23 domains, with an average video duration of 23 minutes. Considering the limitations of existing MLLMs in processing long videos, we segmented the videos into sub-videos, each containing 1-3 slides. This resulted in a total of 74,000 sub-videos, with an average duration of 35 seconds each. Our dataset, compared to existing ones, offers a richer variety in the number and types of videos and QA pairs, features longer video lengths, and introduces previously untouched document scenarios.

\section{Method}
\begin{figure}[t]
    \centering
    \includegraphics[width=1\columnwidth]{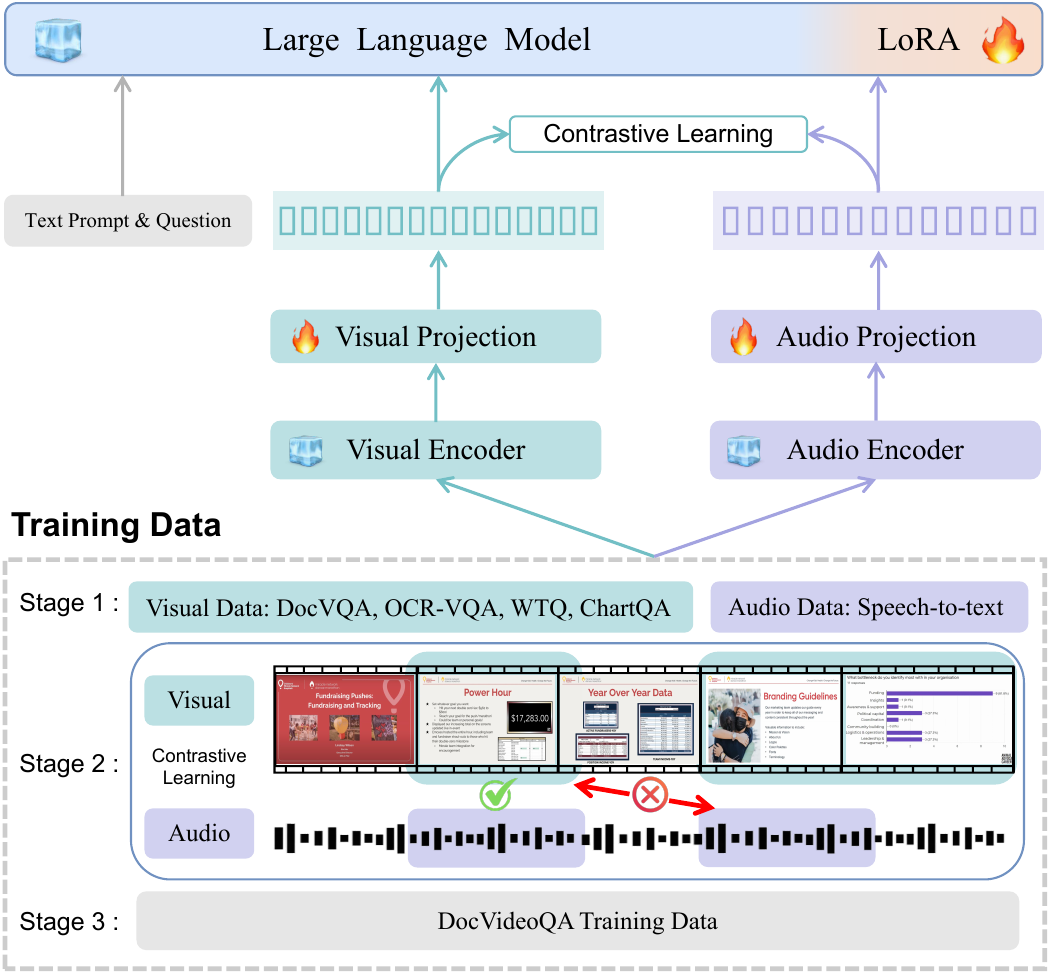}
    \caption{Overview of DV-LLaMA and stage-wise training}
    \label{model_pic}
\end{figure}
\setlength{\textfloatsep}{5pt}
Document videos contain rich audio-visual information, offering detailed explanations of images with text. This requires strong unimodal feature extraction and an understanding of cross-modal relationships for effective integration. As shown in Fig~\ref{model_pic}, inspired by the Video-LLaMA model, we designed our model DV-LLaMA with two branches: Vision-Language and Audio-Language. We also propose a multi-stage training approach, using instruction data and contrastive learning to progressively tackle the challenges of feature extraction and integration in document videos.

\subsection{Single-Branch Feature Enhancement Training}
At this stage, we addressed the model's limitations in processing document videos by constructing instruction data and training each branch separately to improve feature extraction. For the visual branch, given the rich textual content in document images, we leveraged open-source datasets such as OCR-Based DocVQA\cite{b11}, OCR-VQA\cite{b22}, SlideVQA\cite{b23}, and TextVQA\cite{b24}. These datasets not only support visual question answering but also help restore the reading order of text, enhancing the model’s text comprehension. For the audio branch, as audio features are ultimately fed into the LLM for understanding, we focused on Speech-to-Text datasets\cite{b34} to improve transcription accuracy and ensure information completeness. During training, we froze the visual and audio encoders and trained a Q-Former\cite{b32} to compress and project tokens, which were then combined with text tokens for processing by the frozen LLM.

\subsection{Contrastive Learning Alignment Training}
In the second stage, our objective is to align content between visual and audio modalities by learning their interrelations. Given that audio segments in document-centric videos typically serve as explanations or supplements to the visual content, and both modalities share the same contextual semantic meanings, we leverage contrastive learning to facilitate the alignment of representation spaces across modalities.

Assuming a batch size of \(B\) during the training process, a visual segment and its temporally aligned audio segment are considered as a positive pair, while pairs formed with other audio segments within the same batch are treated as negative examples. For each positive pair \( (v, a) \) from the distinct data modalities \(V\) and \(A\), an image \(V_i\) is encoded into \(e_v^i\) by the vision encoder, and an audio \(A_i\) is encoded into \(e_a^i\) by the audio encoder. We select \(B-1\) audio embeddings from the same batch as negative examples. Drawing upon common practices in vision-language pretraining, we compute the image-to-audio contrastive learning loss as follows:
\[
L_{CL}^t = -\sum_{i=1}^{B} \frac{1}{B} \log \left[ \frac{f(e_a^i, e_v^i)}{f(e_a^i, e_v^i) + \sum_{k \neq i} f(e_a^i, e_v^k),} \right]
\]
where \(f(e_a^i, e_v^k)\) measures the cosine similarity between \(e_a^i\) and \(e_v^k\) within a semantic space. During this phase, only the projection layers are trained. By utilizing contrastive learning in this phase, we aim to enable the model to understand the similarities in content that co-occur in the visual and audio modalities, thereby promoting cross-modal integration.

\subsection{Multimodal Fusion Question-Answering Training}
In the third phase, we focused on integrating information from visual and audio modalities and interacting with textual data, aiming to fuse the three modalities for multimodal question-answering tasks. For instruction fine-tuning, on one hand, we utilized the open-source dataset VideoInstruct100K\cite{b29}, which contains approximately 100,000 video clips, each annotated with GPT-generated instruction-response pairs. This dataset enhances models' interactive capabilities through scenarios requiring complex comprehension and response generation from visual cues and verbal instructions. On the other hand, we constructed the training set for DocVideoQA, selecting 1.6K videos with QA pairs that were carefully designed and curated to improve the model’s understanding of interactions between text, visual, and audio modalities. During this process, features extracted from the three modalities were concatenated and fed into the LLM, while the video and audio encoders remained frozen and the projectors were optimized in our multimodal training approach. Unlike previous phases, this phase also involved fine-tuning the LLM using LoRA\cite{b25}, transforming the LLM into a true MLLM. By effectively leveraging synchronized audio-visual data, DV-LLAMA achieved a deeper understanding of multimodal content, thereby improving its performance on the DocVideoQA task.

\section{Experiments}

\subsection{Implementation details}
Our model is based on Video-LLaMA, utilizing LLaMA-2-7B\cite{b33}. For the visual encoder, we consistently use the pre-trained vision component from BLIP-2\cite{b26} without any modifications. This includes a ViT-G/14 architecture from EVA-CLIP\cite{b27}, alongside a pre-trained Q-Former. For the audio-language branch, we incorporate the pre-trained ImageBind\cite{b28} as the audio encoder. No hyperparameter tuning is applied during training. Instead, we empirically set the global batch size to 2,048 and the learning rate to 2e-5 for fine-tuning. The fine-tuning process is conducted for up to 3 epochs in each stage. We perform all fine-tuning on a workstation equipped with 4 NVIDIA Tesla A800 GPUs, with 80 GB of memory.
\subsection{Evaluation Criteria}
We evaluate performance using two common VideoQA metrics: \textbf{Accuracy}. For the test set of size $N$, given a question $q_i$ and its ground-truth answer $y_i$, let the model’s predicted answer be $a_i$, where both $a_i$ and $y_i$ are treated as sets of words. The evaluation criteria are as follows:

\textbf{Accuracy}: This is redefined using the \textit{BERT\_score} to account for semantic similarity between predicted and true answers. Accuracy is calculated as:
\[
\text{Accuracy} = \frac{1}{N} \sum_{i=1}^{N} {1}[\text{BERT\_score}(a_i, y_i) > T]
\]
where the indicator function outputs 1 if the \textit{BERT\_score} exceeds threshold $T$, allowing for more flexibility than exact matching.





\subsection{Experiment Result and Analysis}

\begin{table}[t]
\centering
\caption{Accuracy comparison across different domains at $T=0.8$}
\renewcommand{\arraystretch}{1.2} 
\begin{tabularx}{1\columnwidth}{llccccc}
    \toprule
    \multirow{2}{*}{\#} & \multirow{2}{*}{Model} & \multicolumn{4}{c}{$Acc_{T=0.8}$} \\ \cline{3-6}
    \addlinespace[1mm]
    && Agriculture & Education & Fitness & Total \\
    \midrule
    \rowcolor{gray!20} \multicolumn{6}{c}{\textit{Open-source MLLMs}} \\
    1 & Video-Chatgpt\cite{b29} & 52.58 & 54.47 & 52.64 & 53.23 \\
    2 & VideoChat\cite{b30} & 54.72 & 51.69 & 55.71 & 54.62 \\
    3 & Video-LLaMA\cite{b19} & 60.48 & 62.58 & 64.65 & 61.88 \\
    4 & Video-LLaVA\cite{b31} & 62.37 & 68.46 & 62.12 & 63.92 \\
    \rowcolor{gray!10} \multicolumn{6}{c}{\textit{Our Model}} \\
    5 & DV-LLaMA & 68.64 & 72.69 & 71.76 & 70.51 \\
    6 & \quad - First stage & 63.34 & 63.21 & 65.49 & 65.33 \\
    7 & \quad - Second stage & 64.32 & 67.26 & 66.92 & 66.52 \\
    8 & \quad - Third stage & 65.72 & 66.95 & 64.47 & 66.03 \\
    \bottomrule
\label{tab:result}
\end{tabularx}
\end{table}
\setlength{\textfloatsep}{0pt}

Table~\ref{tab:result} presents a comparative analysis between our model DV-LLaMA and several state-of-the-art multi-modal large language models (MLLMs) across various domains within the DocVideoQA Dataset. As shown from row \#1 to row \#4, open-source models do not perform optimally, primarily due to their inability to handle the specific characteristics of document-centric videos. We then applied a three-stage training process, which resulted in DV-LLaMA, as shown in row \#5. The model achieved a 20\% improvement in performance, significantly surpassing existing open-source models.

Specifically, row \#6 indicates that the first training stage plays a crucial role in improving the model's feature extraction capabilities, enabling it to adapt to the complex nature of document-centric videos and better interpret rich-text images and information-dense audio dialogues. Row \#7 shows that contrastive learning aligns information from both visual and audio modalities, allowing the model to learn how audio complements and explains visual content. Finally, row \#8 highlights that by leveraging LoRA to fine-tune both the LLM and the two modality branches, the model effectively integrates multi-modal information and interacts with text, resulting in better performance on QA tasks.

Overall, the comparative analysis demonstrates that DV-LLaMA, with our progressive training approach, significantly enhances the model's ability to extract and fuse multi-modal information for complex QA tasks involving document videos.

\section{Conclusions}
In this paper, we introduce the task of Document Video Question Answering (DocVideoQA), filling the gap in videoQA for document-centric videos and extending DocumentQA to the most complex modality of video. To further advance research in this area, we provide a corresponding dataset. By developing and evaluating this dataset, we propose DV-LLaMA, which leverages multi-stage fine-tuning and contrastive learning to enhance the understanding and integration of visual and audio features required for document video scenarios in MLLMs. The experimental results powerfully demonstrate the effectiveness of our approach.

\section{Acknowledgement}
This work is supported by the project of Beijing Science and Technology Programme (Z231100007423011), which is also a research achievement of State Key Laboratory of Multimedia Information Processing and Key Laboratory of Science, Technology and Standard in Press Industry (Key Laboratory of Intelligent Press Media Technology).

\end{document}